# AMLID: An Adaptive Multispectral Landmine Identification Dataset for Drone-Based Detection

James E. Gallagher[1*], Edward J. Oughton[1]

[1] Department of Geography & Geoinformation Science, George Mason University, Fairfax, VA 22030, USA

*Corresponding author: jgalla5@gmu.edu*

## Abstract

Landmines remain a persistent humanitarian threat, with an estimated 110 million mines deployed across 60 countries, claiming approximately 26,000 casualties annually. Current detection methods are hazardous, inefficient, and prohibitively expensive. We present the Adaptive Multispectral Landmine Identification Dataset (AMLID), the first open-source dataset combining Red-Green-Blue (RGB) and Long-Wave Infrared (LWIR) imagery for Unmanned Aerial Systems (UAS)-based landmine detection. AMLID comprises of 12,078 labeled images featuring 21 globally deployed landmine types across anti-personnel and anti-tank categories in both metal and plastic compositions. The dataset spans 11 RGB-LWIR fusion levels, four sensor altitudes, two seasonal periods, and three daily illumination conditions. By providing comprehensive multispectral coverage across diverse environmental variables, AMLID enables researchers to develop and benchmark adaptive detection algorithms without requiring access to live ordnance or expensive data collection infrastructure, thereby democratizing humanitarian demining research.

## Background & Summary

Landmines constitute one of the most persistent and devastating post-conflict humanitarian challenges. An estimated 110 million landmines are dispersed across 60 countries, remaining active for decades after hostilities end [1–3]. Each year, approximately 26,000 people are killed or injured by landmines, with children representing 42% of these victims[4]. Beyond the human toll, the economic and social costs are profound, rendering vast tracts of potentially productive land unusable and impeding post-conflict recovery and development [4,5]. Traditional landmine detection methods remain hazardous, inefficient, and prohibitively expensive [6,7]. Using existing demining methods would require decades to clear landmines worldwide. The ongoing conflict in Ukraine has intensified this urgency, with extensive minefields creating an unprecedented humanitarian crisis[8,9].

Recent technological advances offer promising avenues for revolutionizing landmine detection. Unmanned Aerial Systems (UAS) have facilitated detection from safer distances with significantly broader coverage compared to traditional ground-based methods[10,11]. Deep learning models, particularly computer vision algorithms such as the You Only Look Once (YOLO) architecture, have demonstrated remarkable improvements in speed, accuracy, and edge deployment capabilities[12–14]. Furthermore, Long-Wave Infrared (LWIR) sensors excel at detecting surface-laid landmines by capturing distinct thermal signatures created when metallic and plastic objects transfer heat at different rates when compared to the surrounding soil[15–17].

Despite these advances, a critical gap exists in publicly available datasets that combine RGB and LWIR imagery for landmine detection research. Existing datasets are either proprietary, limited to a single spectral modality, or lack the environmental diversity needed to develop robust detection algorithms. This limitation has constrained the broader research community's ability to contribute to humanitarian demining solutions.

To address this gap, we present the Adaptive Multispectral Landmine Identification Dataset (AMLID). AMLID is the first open-source dataset specifically designed to support adaptive RGB-LWIR fusion research for UAS-based landmine detection. The dataset comprises 12,078 labeled images featuring 21 globally-deployed landmine types, spanning both Anti-Personnel (AP) and Anti-Tank (AT) categories in metal and plastic compositions (Figure 1). AMLID provides comprehensive coverage across 11 fusion levels from pure RGB to pure LWIR, four sensor altitudes (5m, 10m, 15m, 20m), two seasonal periods (January and May), and three daily illumination conditions (post-sunrise, noon, pre-sunset).

The design philosophy underlying AMLID recognizes that optimal sensor fusion strategies vary with environmental conditions. As illustrated in Figure 2, conventional binary approaches require operators to select between RGB or LWIR modalities independently. The RGB image provides high spatial resolution but lacks thermal contrast between soil and landmines. In contrast, LWIR captures thermal signatures (appearing as bright spots against cooler soil) but at lower resolution. This treats sensor selection as an either-or decision rather than a continuous optimization problem. However, air-ground temperature differential, time-of-day, illumination intensity, soil composition, and seasonal factors all influence the relative contributions of RGB and LWIR modalities to detection performance. The lower panel of Figure 2 demonstrates how RGB-LWIR fused images from our surface-laid minefield vary along the fusion continuum,

## Landmines used in the AMLID Dataset

Commonly used landmines consisting of a mix of metal and plastic AP and AT mines.

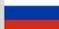

Figure 1. Anti-Personnel (AP) and Anti-Armor (AT) landmines used for this research, consisting of 10x plastic AT, 2x AT metal, 3x AP metal, and 6x AP plastic mines.

with each environmental variable shifting the optimal blend point. By providing imagery across this complete parameter space, AMLID enables researchers to develop adaptive methods that optimize fusion ratios along the RGB-LWIR continuum based on environmental variables, rather than relying on static sensor configurations.

AMLID's potential applications extend beyond algorithm development to include transfer learning benchmarks, sensor fusion methodology validation, and operational planning for humanitarian demining missions. By making this dataset freely available, we aim to democratize access to multispectral landmine detection research and accelerate progress toward freeing affected populations from landmine threats.

## Methods

### Landmine Selection

This research utilized 21 inert landmine simulants provided by the U.S. Army's Counter Explosive Hazards Center (CEHC). The mines were selected to represent the diversity of landmines encountered in humanitarian demining operations and ongoing conflicts. The collection includes mines manufactured in Russia, the United States, Italy, China, Israel, the former Yugoslavia, and Czechoslovakia.

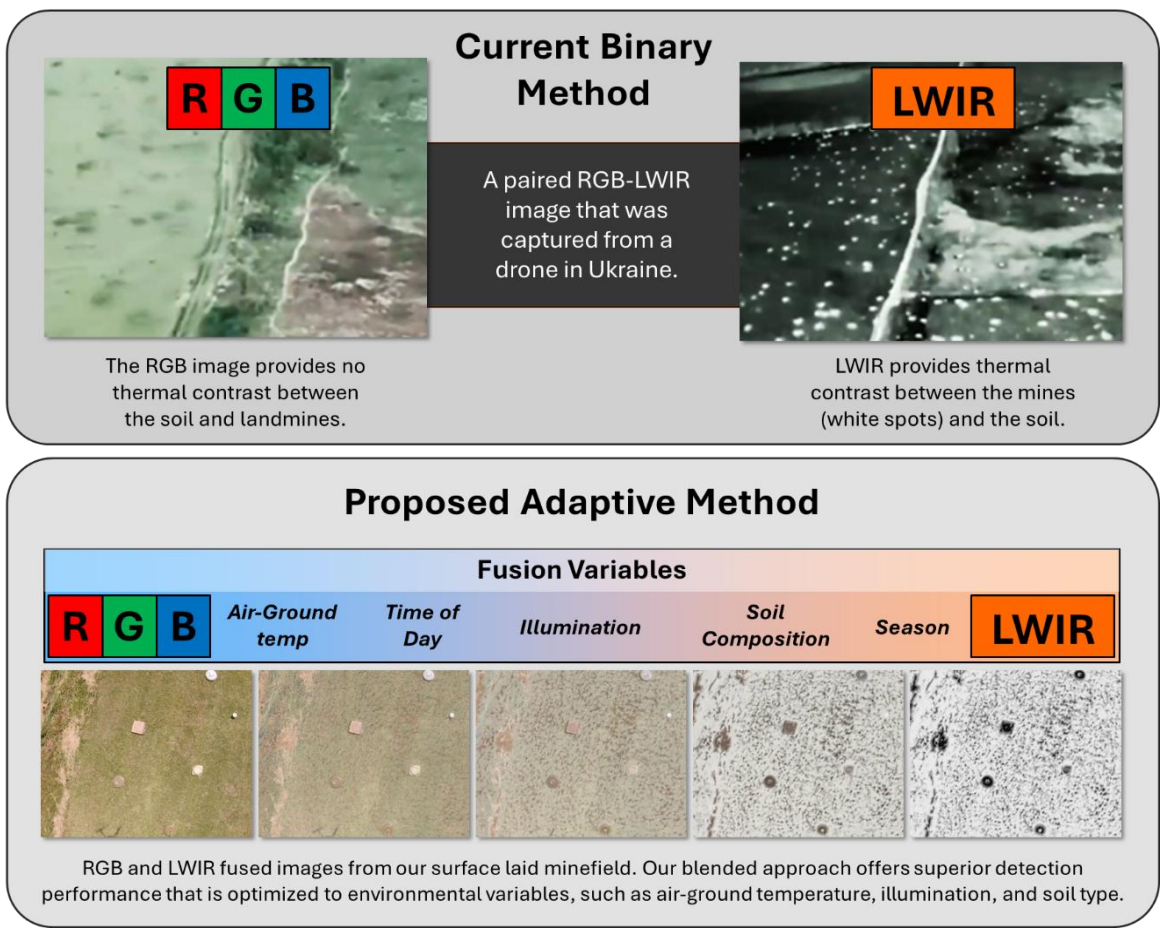

Figure 2. Conventional binary approaches versus the proposed adaptive multispectral method. Current practice requires operators to select between RGB or LWIR modalities independently. The AMLID dataset enables development of adaptive fusion approaches that optimize RGB-LWIR integration along a continuum based on environmental variables including air-ground temperature differential, time of day, illumination, and season.

The mines are categorized along two primary dimensions. First, the dataset consists of Anti-Tank (AT) mines, which are three to six times larger in both dimension and weight since they are designed to penetrate armored vehicles, and Anti-Personnel (AP) mines, which are smaller and designed for use against personnel. Second, material composition differentiates metal mines from plastic mines. Metal mines, while more easily detected with metal-detection systems, are increasingly rare in contemporary conflicts due to their susceptibility to detection and corrosion. Newer mines are predominantly composed of plastic, making them considerably more challenging to detect with metal detection technology.

The dataset comprises of ten plastic AT mines (TM-62, TM-46, TMP-3, TC/2.4, VS-HCT2, SB-81, M19, TC-3.6, TC-6, CV-PT-MI-BA), two metal AT mines (TM-57, MON-100), six plastic AP mines (POMZ-2M, VS-MK2, VS-50, PP MI-NA 1, PMN, APERS NO.4), and three metal AP mines (OZM-72, OZM-3, Type-69). This distribution reflects the prevalence of plastic-cased mines in contemporary minefields while ensuring representation of all major categories.

## Sensor Configuration

The RGB-LWIR fusion method employed to collect the data utilizes feature alignment, with an alpha adjustment applied to the LWIR layer while the RGB layer remains unmodified. The LWIR sensor selected was a FLIR Vue Pro R radiometric camera with a 45 ° field of view (FOV) and a 6.8 mm lens diameter. The LWIR sensor provides a resolution of 336 × 256 pixels with a spectral band of 7.5–13.5 μm, enabling capture of thermal radiation in the long-wave infrared spectrum where landmine thermal signatures are most pronounced.

The RGB sensor selected was the RunCam 5, which uses a Sony IMX377 12-megapixel image sensor with a FOV of 145° and resolution of 1920 × 1080 pixels. A custom 3D-printed housing was designed and fabricated to mount both cameras as close as possible, minimizing parallax between the two imaging modalities. Both sensors were mounted on a DJI Inspire 2 UAS.

## Data Collection Protocol

Data collection was conducted at a controlled site in Leavenworth, Kansas, during January and May to capture seasonal variation in environmental conditions. These months were selected to represent winter and early-summer conditions, with substantial differences in temperature, vegetation state, and solar angle. Table 1 provides detailed environmental measurements for each collection period.

TABLE I
Environmental Variable Measurements During January and May in Leavenworth, Kansas

| Variable | January | | | May | | | Average |
| --- | --- | --- | --- | --- | --- | --- | --- |
| | Post-Sunrise | Noon | Pre-Sunset | Post-Sunrise | Noon | Pre-Sunset | |
| Ground Temp (C°) | -10.9 | 8.5 | 2.8 | 21.4 | 33.5 | 34.4 | 14.9 |
| Air Temp (C°) | 9.9 | 20.7 | 18.9 | 22.6 | 27.4 | 24.7 | 20.7 |
| Lux (lx) | 3,716 | 36,300 | 5,498 | 11,500 | 83,600 | 31,300 | 28,652 |

Table 1. Environmental variables for data collection periods.

The average air temperature for January collections was 16.5°C, while May averaged 29.8°C, resulting in a 13.3°C difference between the seasons. Ground temperature exhibited even greater variation: 0.13°C in January versus 29.76°C in May, a difference of 29.63°C. These temperature differentials create distinct thermal contrast conditions that influence the detectability of landmine thermal signatures.

Within each seasonal period, data was collected during three daily time periods: post-sunrise, noon, and pre-sunset. These times were selected to capture the full range of illumination, solar angle, and temperature conditions. The post-sunrise period exhibited the coldest average ground and air temperatures (5.2°C and 16.3°C, respectively) with low illumination (7,608 lux). The noon period is characterized by high ground and air temperatures (21°C and 24°C) and very high illumination (59,950 lux), with minimal shadows. The pre-sunset period has mid-range temperatures (18.6°C ground, 21.8°C air) and medium illumination (18,399 lux).

Data was collected at four fixed elevations: 5m, 10m, 15m, and 20m. These altitudes span the operational range relevant to UAS-based landmine detection, from close-range, high-resolution imaging to broader-area coverage at higher altitudes with medium- to low-resolution imagery. As illustrated in Figure 3, AMLID's systematic organization across temporal periods, fusion levels, and elevations provides researchers with a comprehensive framework for training and evaluating computer vision models under controlled experimental conditions. The dataset's structure enables systematic investigation of how detection algorithms perform across the multidimensional parameter space encountered in real-world humanitarian demining operations, including seasonal variation, daytime illumination cycles,

## Proposed Method for Utilizing AMLID
Data collection, model training, and sensor evaluation for mines at different altitudes, seasons, and illumination conditions.

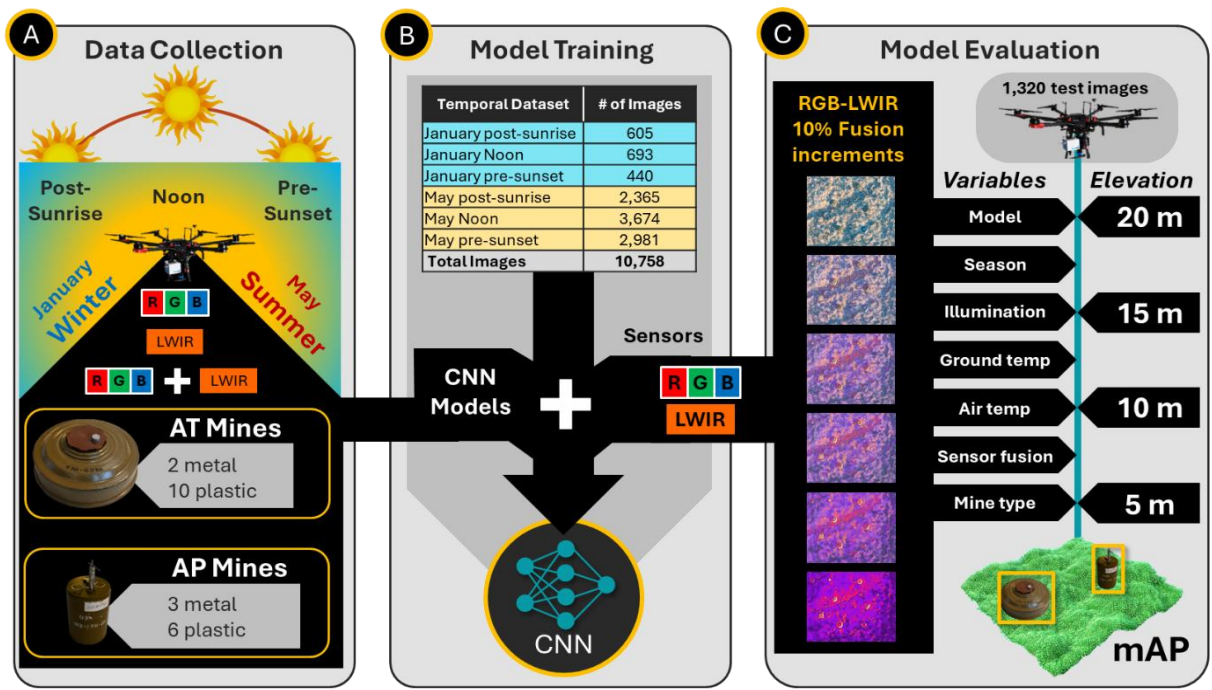

Figure 3. AMLID data collection and organization methodology. (A) Data collection encompassed two seasonal periods (January and May) with three daily time periods (post-sunrise, noon, pre-sunset) using co-registered RGB and LWIR sensors mounted on a DJI Inspire 2 UAS. (B) Model training framework showing the distribution of 10,758 training images across temporal datasets for each of the 27 YOLO model configurations (3 architectures × 9 temporal combinations). (C) Evaluation methodology using 1,320 balanced test images across 11 RGB-LWIR fusion levels (0-100% in 10% increments) and four sensor elevations (5m, 10m, 15m, 20m).

sensor fusion strategies, and altitude-dependent resolution trade-offs. This comprehensive dataset coverage enables the computer vision community to benchmark object detection architectures, develop altitude-adaptive algorithms, and identify optimal operational parameters without requiring access to live ordnance or costly data-collection infrastructure.

## Image Fusion and Processing

Following data collection, RGB and LWIR footage were fused post-flight, with RGB imagery as the base layer and LWIR imagery overlaid using feature alignment. The LWIR alpha level was adjusted in increments of 10%, creating 11 distinct fusion levels: 0% (pure RGB), 10%, 20%, 30%, 40%, 50%, 60%, 70%, 80%, 90%, and 100% (pure LWIR). This systematic approach enables researchers to investigate the full spectrum of RGB-LWIR integration strategies.

Video frames were extracted every 120 frames (2-second intervals) for each fusion level. All images underwent manual quality control to remove blurry or degraded frames. Following the quality review, all landmine instances were manually annotated. Due to the uncommon nature of landmines as an object class, automated labeling techniques using standard datasets such as COCO yielded unsuccessful results, necessitating comprehensive manual annotation [18].

## Annotation Protocol

All landmine instances were annotated using bounding boxes with class labels corresponding to the four mine categories: AP-metal, AP-plastic, AT-metal, and AT-plastic. Annotations were created in Pascal VOC XML format and subsequently converted to YOLO format using normalized bounding box coordinates. Each annotation includes the object class and bounding box coordinates (center x, center y, width, height) normalized to image dimensions. Quality control procedures included verification of bounding box accuracy by a second annotator on a random 10% sample for consistency checks, to ensure that all visible mines in each frame were labeled.

## Data Records

The Adaptive Multispectral Landmine Identification Dataset (AMLID) is available for download from Zenodo[19].

## Dataset Structure

The complete dataset comprises 12,078 images organized according to the following hierarchy: seasonal period (January/May), time-of-day (post-sunrise/noon/pre-sunset), and fusion level (0-100% in 10% increments). The dataset is divided into training (10,758 images, 89%) and testing (1,320 images, 11%) subsets.

The temporal distribution reflects constraints on operational data collection. January collections account for 1,738 training images (16.2%) distributed across post-sunrise (605 images), noon (693 images), and pre-sunset (440 images) periods. May collections comprise 9,020 training images (83.8%) distributed across post-sunrise (2,365 images), noon (3,674 images), and pre-sunset (2,981 images) periods.

The testing dataset maintains balanced representation across operational conditions, with each of the six temporal periods contributing 220 images distributed equally across four elevation levels (5m, 10m, 15m, 20m), yielding 55 images per elevation/temporal combination.

## File Formats

Images are provided in JPEG format at native sensor resolution. Each image is accompanied by a corresponding Pascal VOC XML annotation file containing bounding box coordinates and object class.

## Ground Truth Statistics

The test dataset contains 14,905 labeled landmine instances. AT mines account for 79.2% of total labels, while plastic-cased mines constitute 79.3% of all mines. The specific distribution includes AT-plastic (10,032 annotations, 67.3%), AT-metal (1,771 annotations, 11.9%), AP-plastic (1,782 annotations, 12.0%), and AP-metal (1,320 annotations, 8.9%). This distribution reflects both the physical characteristics of the mine collection and the relative visibility of different mine types across elevation and fusion conditions.

## Technical Validation

### Sensor Calibration

The FLIR Vue Pro R radiometric camera underwent factory calibration before data collection. Thermal calibration was verified using blackbody reference sources at known temperatures. RGB camera exposure settings were standardized across collection periods to ensure consistent color reproduction.

### Spatial Registration

RGB-LWIR image pairs were spatially registered using feature-based alignment to compensate for differences in sensor field of view and mounting position. Registration accuracy was validated by visual inspection of fused imagery to confirm alignment of standard features (mine boundaries, ground markers) across the spectral modalities. The custom 3D-printed camera mount minimized baseline separation between sensors, reducing parallax effects across various altitudes.

### Annotation Quality

Annotation quality was ensured through multiple validation steps. Initial annotations were created by trained personnel familiar with the visual characteristics of landmines. A random 10% sample underwent independent verification by a second annotator.

### Environmental Measurement Verification

Air temperature, ground temperature, and illumination measurements were recorded using calibrated instruments. Air temperature was measured using a certified thermometer positioned at a height of 1.5m in the shade. Ground temperature was measured at multiple locations within the test area using an infrared thermometer. Illumination was measured using a calibrated lux meter. Measurements were recorded at the beginning and end of each collection period, with reported values representing period averages (Table 1).

### Landmine Simulant Verification

All landmine simulants were provided by the U.S. Army Counter Explosive Hazards Center and verified to match the external dimensions, material composition, and surface characteristics of their live counterparts. While simulants lack internal explosive components, their thermal and visual signatures are representative of surface-laid ordnance. All landmine simulants in AMLID are surface laid.

## Usage Notes

AMLID is designed to support a range of computer vision and machine learning applications for landmine detection research. Potential use cases include training and benchmarking object detection models (such as YOLO, Faster R-CNN, and transformer-based architectures), investigating optimal RGB-LWIR fusion strategies across environmental conditions, developing adaptive algorithms that dynamically adjust sensor fusion based on environmental variables, and transfer learning from this domain to other related remote sensing applications.

Researchers should note the class imbalance between AT and AP mine categories. Standard techniques for handling class imbalance (weighted loss functions, oversampling, data augmentation) may be appropriate depending on research objectives. The temporal imbalance between January (16.2%) and May (83.8%) training data should also be considered when developing season-specific models. Researchers investigating seasonal generalization may wish to balance these proportions through sampling or augmentation. Lastly, all data was collected at a single geographic site with relatively homogeneous soil properties. Models trained on AMLID may require fine-tuning or domain adaptation for deployment in regions with substantially different terrain, vegetation, or soil characteristics.

## Data Availability

The Adaptive Multispectral Landmine Identification Dataset (AMLID) is publicly available at Zenodo[19]. The dataset includes all 12,078 images with corresponding Pascal VOC annotations.

## Code Availability

No code was used to produce the AMLID dataset.

## Acknowledgements

The authors thank the U.S. Army Counter Explosive Hazards Center (CEHC) for providing us access to the landmine simulants.

## Author Contributions

J.E.G. designed the study, collected and processed data, performed annotations, conducted analysis, and wrote the manuscript. E.J.O. supervised the research, contributed to the study design, and revised the manuscript.

## Funding

No funding was used for this research.

## Competing Interests

The authors declare no competing interests.